\pdfoutput=1
\documentclass[11pt]{article}

\usepackage[final]{acl}

\usepackage{times}
\usepackage{latexsym}

\usepackage[T1]{fontenc}
\usepackage[utf8]{inputenc}

\usepackage{microtype}

\usepackage{inconsolata}

\usepackage{graphicx}
\usepackage{float}

\usepackage{times}
\usepackage{helvet}
\usepackage{courier}
\usepackage{amsmath}
\usepackage{graphicx}
\usepackage{subcaption}
\usepackage{caption}
\usepackage{hyperref}
\usepackage{bm} 

\usepackage{algorithm}
\usepackage[noend]{algpseudocode}

\usepackage{booktabs}

\usepackage{color}

\title{STAR: Spectral Truncation and Rescale for Model Merging}

\author{\textbf{Yu-Ang Lee\textsuperscript{1, 2}}\thanks{This work was done while Yu-Ang Lee was a visiting
researcher at IBM Thomas J. Watson Research Center.},
 \textbf{Ching-Yun Ko\textsuperscript{2}},
 \textbf{Tejaswini Pedapati\textsuperscript{2}},
 \\
 \textbf{I-Hsin Chung\textsuperscript{2}},
 \textbf{Mi-Yen Yeh\textsuperscript{3}},
 \textbf{Pin-Yu Chen\textsuperscript{2}}
\\
\\
\textsuperscript{1}National Taiwan University,
\textsuperscript{2}IBM Research,
\textsuperscript{3}Academia Sinica
\\
\texttt{r12946015@ntu.edu.tw, cyko@ibm.com, tejaswinip@us.ibm.com} \\
\texttt{ihchung@us.ibm.com, miyen@iis.sinica.edu.tw, pin-yu.chen@ibm.com}
}

\begin{document}
\maketitle
\begin{abstract}
Model merging is an efficient way of obtaining a multi-task model from several pretrained models without further fine-tuning, and it has gained attention in various domains, including natural language processing (NLP). Despite the efficiency, a key challenge in model merging is the seemingly inevitable decrease in task performance as the number of models increases. In this paper, we propose \textbf{S}pectral \textbf{T}runcation \textbf{A}nd \textbf{R}escale (STAR) that aims at mitigating ``merging conflicts'' by truncating small components in the respective spectral spaces, which is followed by an automatic parameter rescaling scheme to retain the nuclear norm of the original matrix. STAR requires no additional inference on original training data and is robust to hyperparamater choice. We demonstrate the effectiveness of STAR through extensive model merging cases on diverse NLP tasks. Specifically, STAR works robustly across varying model sizes, and can outperform baselines by 4.2\% when merging 12 models on Flan-T5. Our code is publicly available at \href{https://github.com/IBM/STAR}{https://github.com/IBM/STAR}.

\end{abstract}

\section{Introduction}
\begin{figure}[t]
    \centering
    \includegraphics[width=0.9\linewidth, height=4cm]{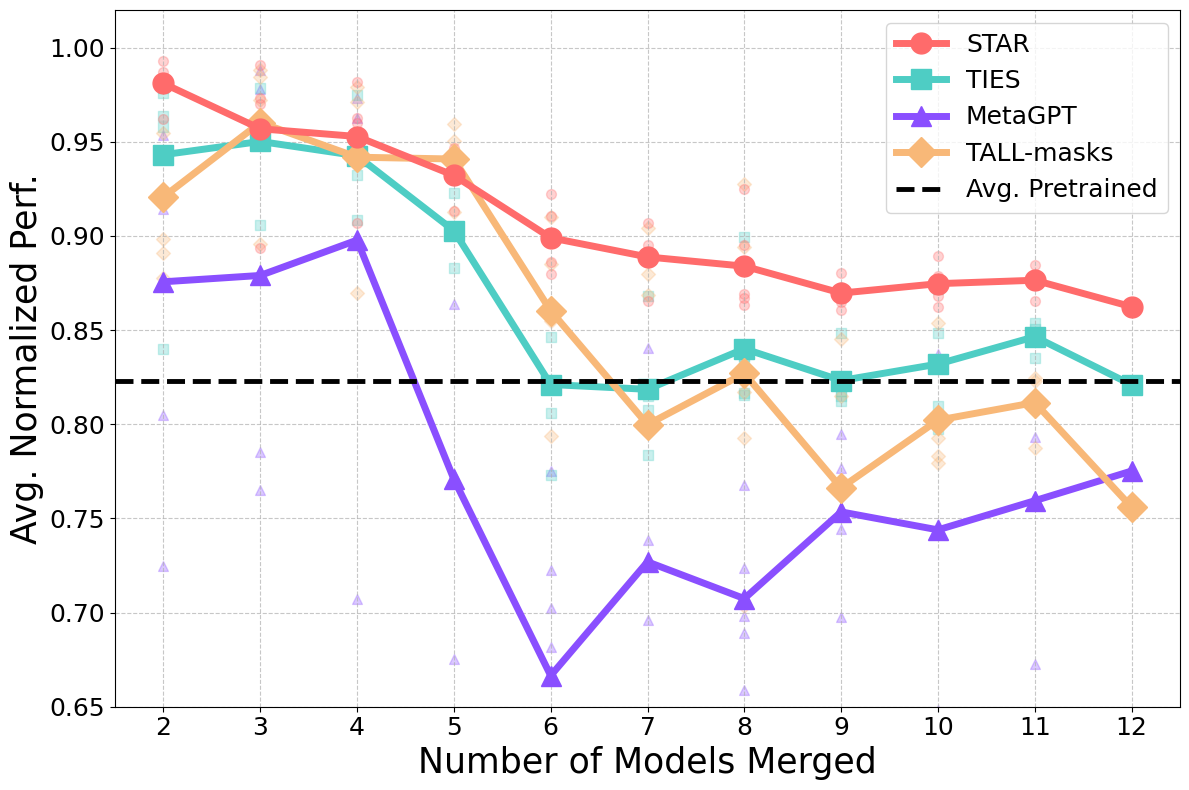}
    \caption{The averaged normalized performance of Flan-T5-base merged models by TIES~\cite{yadav2024ties}, MetaGPT~\cite{zhou2024metagpt}, TALL-masks~\cite{wang2024localizing}, and STAR (this paper).} 
    \vspace{-1em}
    \label{fig:numerial_evidence}
\end{figure}

With the popularity of pretrained models on large neural networks, the same architecture is often deployed to fine-tune individual natural language processing (NLP) tasks. A natural question then arises about whether it is possible to merge these same-architecture fine-tuned models into one multi-task model. For example, researchers are interested in understanding if we can empower a fine-tuned conversational large language model (LLM) with reasoning capabilities by merging with an LLM specializing in solving math problems.  Specifically, ~\citet{ilharco2022editing} has formally defined a \emph{task vector} as \(\theta_{\text{ft}} - \theta_{\text{pre}}\), where $\theta_{\text{pre}}$ and $\theta_{\text{ft}}$ denote the vectorized parameters of the pre-trained model and the fine-tuned model, respectively. Thus, task vectors mark the updates made to the pretrained model's weights when fine-tuned on specific tasks. Then, \textit{model merging} essentially studies ways of fusing different task vectors that are trained separately and merging them with the pretrained model. However, as the number of fine-tuned models increases, the multi-task performance of their merged model also decreases drastically. Fig.~\ref{fig:numerial_evidence}
shows the averaged normalized performance (y-axis) v.s.  the number of models merged (x-axis).
%the x-axis denotes the number of models merged and the y-axis denotes the averaged normalized performance.
% (i.e. average normalized accuracy). 
Furthermore, we point out that when the number of models exceeds a certain threshold, the multi-task performance of the merged model could be even worse than that of the original pretrained model, diminishing the fundamental goal of model merging. For example, TIES~\cite{yadav2024ties}, MetaGPT~\cite{zhou2024metagpt}, and TALL-masks~\cite{wang2024localizing} merged models drop below 0.82 when we merge 6, 5, and 7 fine-tuned models, respectively, in Fig.~\ref{fig:numerial_evidence}. 

\begin{figure*}[t]
    \centering
    \vspace{-1em}
    \includegraphics[width=0.85\linewidth, height=5.cm]{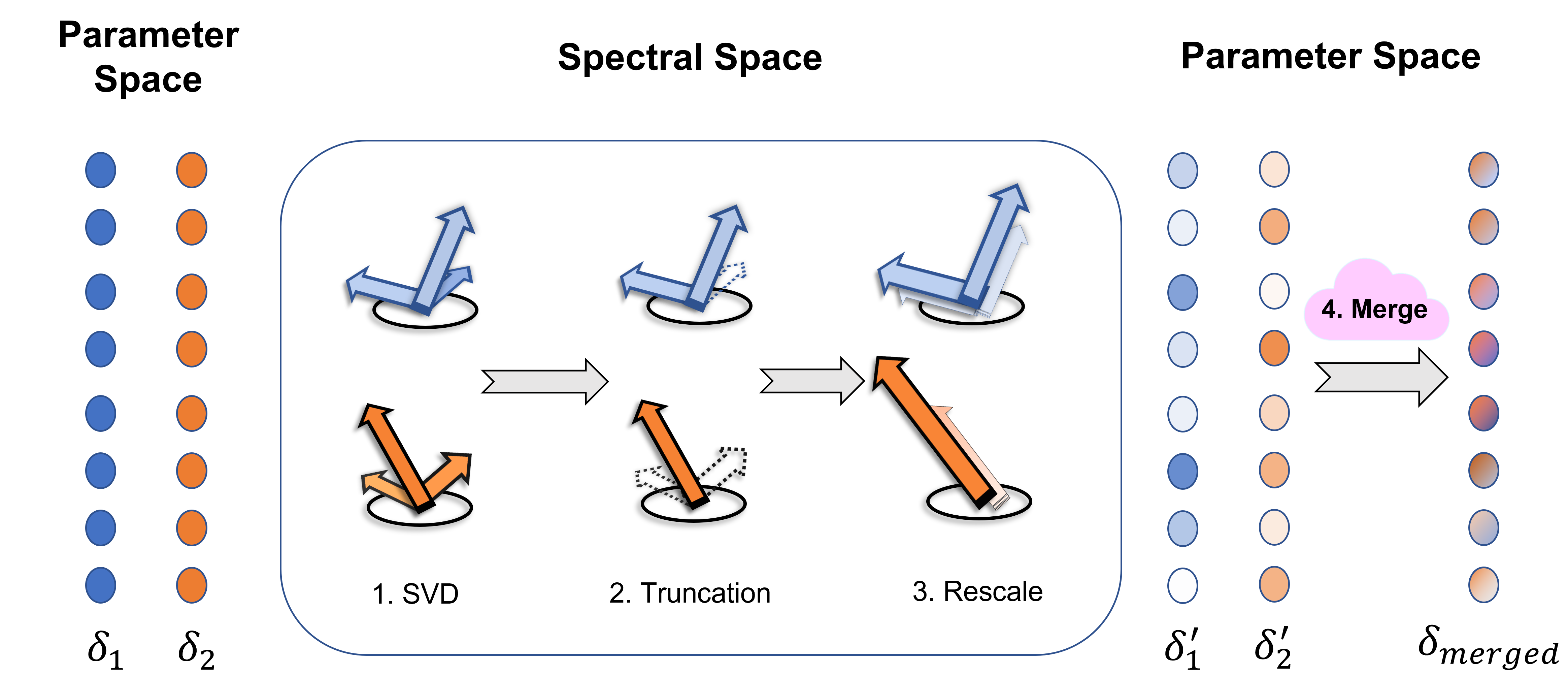}
    \caption{An overview of the \textbf{STAR workflow}. When merging two task vectors, \(\delta_{1}\) and \(\delta_{2}\), (1) STAR transforms both task vectors into their spectral spaces with their singular vectors being the orthogonal basis using singular value decomposition (SVD) (singular values are represented by the length of the arrows), (2) STAR removes redundant dimensions by truncating singular vectors with small singular values, (3) STAR restores the original nuclear norm by rescaling the truncated SVD, and (4) STAR reconstructs the parameters by multiplying components back to form the weight matrices and then perform simple averaging.}
    \vspace{-1em}
    \label{fig:fig1}
\end{figure*}
The complexity of existing model merging methods varies largely depending on whether they require fine-tuning or inference on training data~\cite{yang2024model}. In this paper, we study the ``data-free'' setting when we are not authorized to change the fine-tuning protocol nor do we have access to the training data. In this work, we propose to use spectral decomposition (e.g. singular value decomposition, SVD) to remove noisy components on model merging. We will also motivate the potential gain of our spectral space merging scheme by comparing the upper bounds of the task conflicts. A rescaling step is then followed to restore the original nuclear norm. We give the overview of the proposed method in Fig.~\ref{fig:fig1}. Our proposed merging scheme, \textbf{S}pectral \textbf{T}runcation \textbf{A}nd \textbf{R}escale (STAR), is effective and efficient as it requires no additional inference on original training data and is not sensitive to hyperparameters.
Our extensive experimental results show that STAR is superior across various model size settings and can effectively merge up to 20 models while achieving positive performance gains, compared to the pretrained model before merging.

\section{Background and Related Work}
\label{sec:background}

\subsection{Notations and Problem Definition}\label{subsec:problem_def}

% \textbf{Task Vectors.} 
We denote the weight matrices of a pretrained LM by \(\bm{\theta}_{\text{pre}}^{l}\) for \(l = \{ 1, \ldots, L\} \), where \(L\) is the total number of such matrices. Let $\bm{\theta}_{\text{pre}}$ denote the concatenation of all vectorized weight matrices and $\bm{\theta}_{\text{ft}}$ denote the updated model parameters after fine-tuning on task $\mathcal{T}$. 
A task vector \(\bm{\delta}\) is then defined as the difference between \(\bm{\theta}_{\text{ft}}\) and \(\bm{\theta}_{\text{pre}}\), i.e., \(\bm{\delta} = \bm{\theta}_{\text{ft}} - \bm{\theta}_{\text{pre}}\)~\cite{ilharco2022editing}.
% \noindent\textbf{Model Merging.} 
Given $T$ fine-tuned models, model merging fuses $\{\bm{\delta}_1,\ldots,\bm{\delta}_T\}$ into a merged $\bm{\delta}_{\text{merged}}$ such that $\bm{\theta}_{\text{pre}}+\bm{\delta}_{\text{merged}}$ still performs well on $T$ tasks simultaneously.

\subsection{Related Work}

% \textbf{Model Merging.} 
Model merging methods belong to two categories: Pre-merging and During-merging methods~\cite{yang2024model}. While pre-merging methods focus on renovating the fine-tuning step such that the fine-tuned models suit model merging better~\cite{ortiz2024task,imfeld2023transformer,guerrero2022re}, during-merging methods assume no access to the fine-tuning and work directly on models given. Recently, \citet{yang2024model} further classifies during-merging methods into five sub-classes,
% basic, weighted-based, subspace-based, routing-based, and post-calibration methods, 
of which STAR is most related to the weighted-based and subspace-based methods.

\noindent\textbf{Weighted-based.}
As base merging methods such as~\citet{ilharco2022editing} applies the same scaling across all model layers and tasks, weighted-based methods take the importance of parameters into account and scale differently, e.g. \citet{matena2022merging,tam2024merging} leverage Fisher matrix for assessing the importance of parameters, while others utilize Hessian estimation or entropy, etc~\cite{daheim2023model,yang2023adamerging}. 
% , which could not consider the degrees of importance for different models and layers~\cite{yang2024model}. To bridge this gap,
% Fisher Merging~\cite{matena2022merging}, MaTS~\cite{tam2024merging} propose to leverage Fisher matrix for assessing importance of parameters, while~\citet{daheim2023model} propose uncertainty-based algorithm which merge models based on second-order Hessian estimation. 
% Also, Adamerging~\cite{yang2023adamerging} aims to autonomously learn the coefficient from entropy loss of unlabeled test dataset. 
However, these methods require inference through original data, making it infeasible with limited compute or access to task data.
% encountering limitation when resource is limited or task data is unavailable. 
% To obey the data-free setting commonly in real world, 
MetaGPT~\cite{zhou2024metagpt} proposes a closed form solution for scaling task vectors by minimizing the average loss of the merged model and the independent model.  

\noindent\textbf{Subspace-Based.} Another line of work transforms task vectors into sparse subspaces~\cite{davari2023model,yadav2024ties,wang2024localizing, huang2024emr},
% \IKB{cite the other two papers}
e.g. TIES~\cite{yadav2024ties} trims task vectors to keep only the top \(K\%\) parameters with the highest magnitude, before undergoing an elect-sign step to reduce sign conflicts; TALL-masks~\cite{wang2024localizing} constructs per-task masks that identifies important parameters within each task, which are then merged into one general mask based on consensus among multiple per-task masks.
% Model Breadcrumbs~\cite{davari2023model} instead trims to keep parameters in the middle (not too large nor small) for each weight matrices separately.
% and allows different scaling for . 
% magnitude suggests that essential information lies in the center of the magnitude distribution, and removing both top and bottom outlier parameters could improve merging.

STAR differs from the above as it transforms task vectors to the spectral spaces, and its truncation and scale are task-dependent and layer-specific.

\section{Methodology}\label{sec:spectral_truncate}
Sec. \ref{subsec:truncate} provides the rationale behind performing truncations in the spectral space. Sec. \ref{subsec:rescale} defines the rescaling step for restoring the nuclear norm. Sec. \ref{subsec:STAR} gives the complete STAR algorithm.

\subsection{Spectral Truncation}
\label{subsec:truncate}
Let $\mathcal{T}_1,\mathcal{T}_2$ be two fine-tuning tasks that yield task vectors $\delta_{T_1}$ and $\delta_{T_2}$. Take the entries correspond to a weight matrix and reconstruct them into $A,B$ from $\delta_{T_1}$ and $\delta_{T_2}$, respectively. Suppose $A$ and $B$ admit SVD into $\sum_i\sigma_i^A u_i^A (v_i^A)^T$ and $\sum_i\sigma_i^B u_i^B (v_i^B)^T$, one can obtain the matrix rank by the number of nonzero singular values. By selecting only the top few singular values and vectors (i.e. truncated SVD), we naturally find the principal components and remove the redundant dimensions, effectively reducing the rank of the matrix. As small singular values often correlate with noise or fine details, low-rank prior is also widely used in compressed sensing and denoising applications in signal processing~\cite{dabov2007image,candes2010matrix,cai2010singular,candes2012exact}.

Besides extracting principal components, we also give a high-level illustration of why using truncated SVD on $A$ and $B$ separately can help reduce conflicts during model merging. Assume $\mathcal{T}_1$ is associated with data manifold $\mathcal{D_A}$. For $x\in\mathcal{D_A}$, we essentially hope $(A\oplus B)x$ to be close to $Ax$ while excelling at $\mathcal{T}_2$ after merging, where $\oplus$ denotes the merging operation. Let us consider the merging operation to be plainly $A+B$, then the level of conflicts can be measured by $\|Bx\|$. By expressing $x\in\mathcal{D_A}$ via the right singular vectors of $A$, $x=\sum_j \alpha_j v_j^A$, we prove in Sec.~\ref{subsec:bound} that we have $\|Bx\|\leq r^B\beta \sqrt{r^A}$,
% write out $\|Bx\|$ as  
% % \begin{align*}
% %     &\|\sum_i\sigma_i^B u_i^B (v_i^B)^T \sum_j \alpha_j v_j^A\| \\
% %     \leq &\sum_i\sum_j |\sigma_i^B\alpha_j|\cdot |(v_i^B)^Tv_j^A|\leq \sum_i\sum_j |\sigma_i^B\alpha_j|.
% % \end{align*}
% % \IK{
% {\small
% \begin{align*}
%     &\|\sum_i\sigma_i^B u_i^B (v_i^B)^T \sum_j \alpha_j v_j^A\| \\
%     \leq &\sum_i \|u_i^B\| \cdot |\sum_j \sigma_i^B\alpha_j  (v_i^B)^T   v_j^A| 
%     \leq \sum_i \beta\cdot |\sum_j (v_i^B)^T   v_j^A| \\
%     \leq & \sum_i \beta \sqrt{D^A}\cdot \left(\sum_j \left(\left(v_i^B\right)^T   v_j^A\right)^2\right)^{1/2} \leq r^B \beta \sqrt{r^A},
% \end{align*}}%
where $\beta=\max_{i,j}|\sigma_i^B\alpha_j|$, and $r^A$ and $r^B$ are the original ranks of $A$ and $B$.
% }
By truncating $B$ to rank-$r$, this upper bound is lowered by $(r^B-r)\beta \sqrt{r^A}$
% $\sum_{i=r+1}\sum_j |\sigma_i^B\alpha_j|$
, implying potentially less conflicts in model merging. 
% We denote the task vectors after spectral truncations as $\tilde{\delta}$.
% \begin{figure}
%     \centering
%     \includegraphics[width=1\linewidth]{image/svd_truncate/produce_more_ranks_3.png}
%     \caption{Performing knowledge extraction on \(\delta^{l}\) which is originally with rank \(16\). STAR (\(r=4\)) naturally reduce the total ranks, while TIES (\(K=20\%\)) and Model Breadcrumbs (\(\beta=90\%, \gamma=99\%\)) distort the original basis and introduce more spectral components, which do not resolve task conflicts in spectral perspective.}
%     \label{fig:produce_more_ranks}
% \end{figure}

\subsection{Rescale to Restore Matrix Nuclear Norm}
\label{subsec:rescale}
As model merging favors spectral truncation as discussed in Sec.~\ref{subsec:truncate}, a caveat is the resulting change in the ratio between the pretrained model and the task vector. Roughly, one sees that $\|Ax\|=\|\sum_i\sigma_i^A u_i^A (v_i^A)^T \sum_j \alpha_j v_j^A\|=\|\sum_i\sigma_i^A\alpha_i u_i^A\|$ and can at most be $\sum_{i=r+1}\|\sigma_i^A\alpha_i\|$ smaller with the truncated $A$. Therefore, the performance on the fine-tuning task $\mathcal{T}_1$ might be compromised. 
On that account, it is crucial to include a step where we rescale the spectral-truncated weight matrices back to their original ``size'', similar to the compensation operation in dropout. We propose to retain matrix nuclear norm (aka Schatten $1$-norm or trace norm) as it is a proper measure of matrix ``size'', especially in low-rank approximation contexts as nuclear norm is a convex relaxation of the rank function~\cite{candes2012exact}.
Specifically, we rescale the remaining singular values by
\[
\sigma_k' = \frac{\sum_i \sigma_i}{\sum_{i=1}^r \sigma_i} \cdot \sigma_k, \quad \forall k \in [1, r].
\]

\newcommand{\ra}[1]{\renewcommand{\arraystretch}{#1}}

\subsection{STAR: Spectral Truncate And Rescale}\label{subsec:STAR}

Now that we have elaborated on the two key components of STAR, we explain the complete workflow in the following. With $T$ task vectors, we transform them into respective spectral spaces via SVD, and their ranks are determined by $r = \mathop{\arg\min}_{k} \left( \frac{\sum_{i=1}^k \sigma_i}{\sum_i \sigma_i} \geq \eta\% \right)$, where $\eta$ is a tunable parameter. Then, we follow Section~\ref{subsec:rescale} to rescale back to their original nuclear norm. Finally, STAR reconstructs $T$ task vectors from their decompositions and perform simple averaging to obtain $\bm{\delta}_{\text{merged}}$. We give the full STAR model merging algorithm in Alg.~\ref{alg:STAR} in appendix.

\begin{figure}[t]    
    \centering
    \includegraphics[width=0.52\textwidth, height=3.8cm]{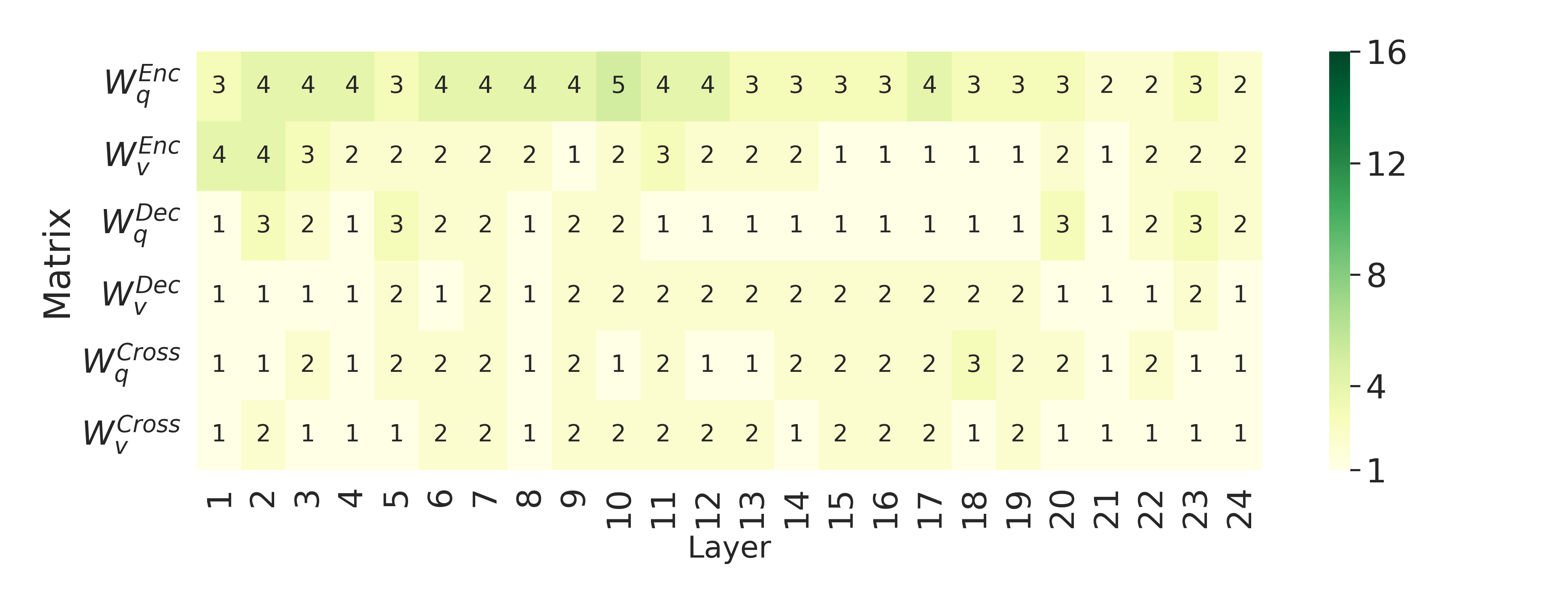}
    \caption{An example of the automatic rank determination by STAR ($\eta=40$) on PIQA's task vector with Flan-T5-large.}
    \label{fig:rank_viz}
\end{figure}

We note that as the distribution of singular values varies both within and across task vectors, truncating components adaptively allows different ranks across not only tasks and even layers (e.g. Fig.~\ref{fig:rank_viz}). 

\section{Experiments}

\begin{figure*}[t]
\vspace{-0.5em}
\begin{subfigure}{0.48\textwidth}
\includegraphics[width=0.99\linewidth]{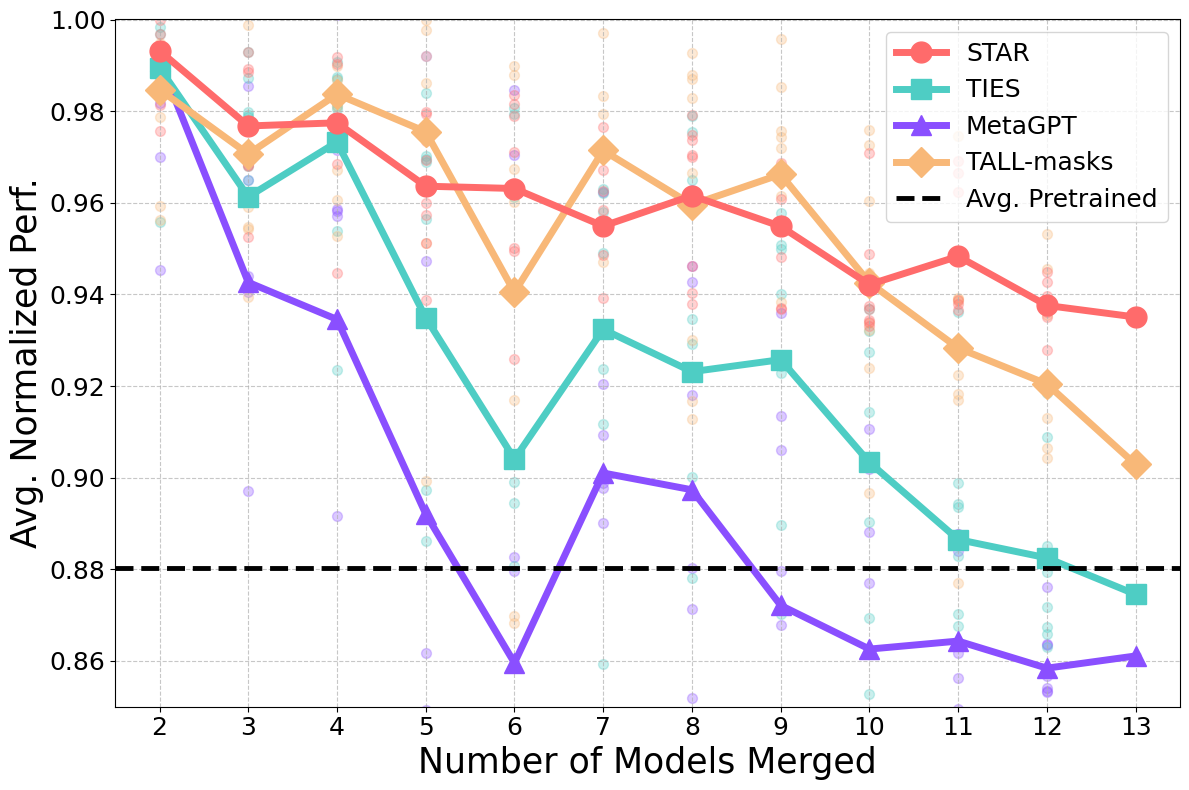} 
\caption{Flan-T5-large}
\end{subfigure}
% \begin{subfigure}{0.33\textwidth}
% \includegraphics[width=0.99\linewidth]{image/main_results/No_tuning/flan_t5_large_no_tune_small_spot.png}
% \caption{Flan-T5-large}
% \end{subfigure}
\begin{subfigure}{0.48\textwidth}
\includegraphics[width=0.99\linewidth]{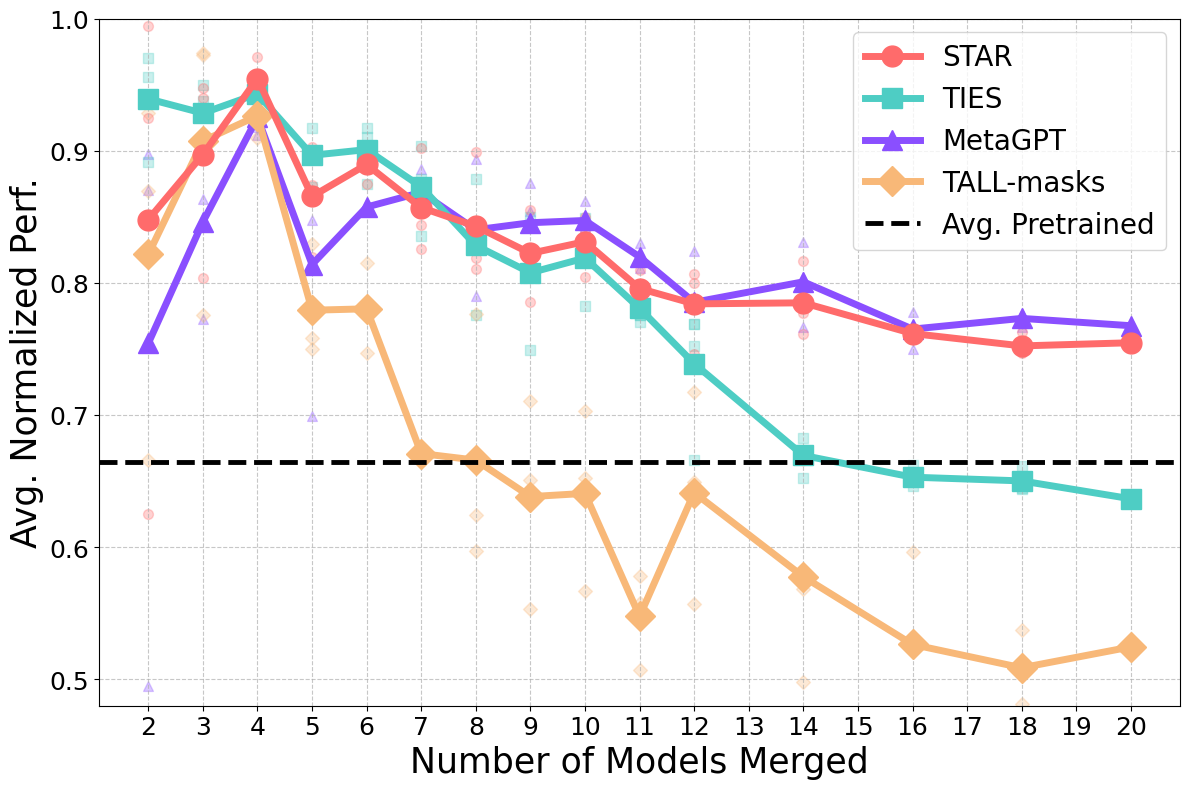}
\caption{Mistral-7B-Instruct}
\label{fig:main_mist}
\end{subfigure}
\caption{Model merging results on Flan-T5-large and Mistral-7B-Instruct. For all numbers of models merged, we sampled 5 task combinations for Flan-T5 and 3 for Mistral, with the sampled combinations represented by shaded dots and the average depicted by solid lines. While STAR remains a strong model merging method, TIES, TALL-masks and MetaGPT can be more sensitive to model architecture choice.}
%\PYB{If all methods are data-free; no need to mention; we can just say in Sec. 2/3 that we consider data-free model merging settings throughout this paper}
\label{fig:main_datafree}
\vspace{-0.5em}
\end{figure*}

% \begin{figure}
%     \centering
%     \includegraphics[width=0.98\linewidth]{image/main_results/No_tuning/mistral_no_tune_small_dots.png}
%     \caption{\textbf{One Shot Merging on Mistral-Instruct.} Up to 20 models were merged in this case, with MetaGPT and STAR show potential to merge even more models without knowledge collapse.}
%     \label{fig:main_mist}
% \end{figure}

\subsection{Experimental Setup}

% \textbf{Pre-Trained Backbone Selection.} 
\noindent\textbf{Models.} We consider both encoder-decoder models (e.g. Flan-T5-base/large)~\cite{chung2024scaling} and decoder-only model (e.g. Mistral-7B-Instruct-v0.2)~\cite{jiang2023mistral}.
For Flan-T5-base/large, we use finetuned models on GLUE from FusionBench~\cite{tang2024fusionbench}, 
% \TP{specify the datasets on which these models were trained on. Also include the model ids for both lots of loras and this in the appendix so that others can reproduce it.} 
together with additional fine-tuned models on Finance~\cite{Malo2014GoodDO}, IMDB~\cite{maas-EtAl:2011:ACL-HLT2011}, AG News~\cite{zhang2015character}, BoolQ~\cite{clark2019boolq}, PIQA~\cite{Bisk2020}, and HellaSwag~\cite{zellers2019hellaswag} by ourselves, bringing the total number of task vectors to \(13\). For Mistral-Instruct, we randomly select \(20\) models directly from the Lots of LoRAs collection~\cite{bruel2024compress}, which covers a range of NLI tasks.
% , which offers a large set of models finetuned on various NLI tasks.
% \(12\) and 
All models considered herein are LoRA finetuned~\cite{hu2021lora} with rank \(16\) and scaling factor (alpha) set to \(32\). Details about the models are in Appendix Sec.~\ref{subsec:experiment_details}.
To understand how each merging method performs on $n$ models, we randomly sample $n$ tasks and report their average results. 

\noindent\textbf{Hyperparameters.} Without otherwise specified, we let $K=20$ for TIES (the default parameter in~\cite{yadav2024ties}), \(\lambda_{t}=0.4\) for TALL-masks (the middle value searched by~\cite{wang2024localizing}), and \(\eta=40\) for STAR. 
% We defer the sensitivity analysis of $K$ and \(\eta\) to the appendix.

\noindent\textbf{Evaluation metric.} Following~\citet{tang2024fusionbench,bruel2024compress}, performances on QASC~\cite{allenai:qasc} and STSB~\cite{cer2017semeval} are evaluated by F1 score and Spearman's coefficient, respectively,
% When evaluating the performance of merged models on individual tasks, F1 score is applied for QASC, Spearman's rank correlation for STSB, 
and accuracy for all other tasks. 
% Despite all three LMs being instruction-tuned, the merged model sometimes outputs answers mixed with random tokens, especially when task conflicts are severe. To account for this, i
If the correct output appears within the first 10 tokens generated by the merged model, the response is deemed correct.
For a model merged on \(t\) tasks, we report the normalized average performance~\cite{ilharco2022editing,yadav2024ties} defined by $\frac{1}{t} \sum_{i}^{t} \frac{\text{(Merged Model Perf.)}_{i}}{\text{(Finetuned Model Perf.)}_{i}}$. We further measure the performance of the pretrained model by $\frac{1}{T} \sum_{i=1}^{T} \frac{\text{Pretrained Model Perf.}_{i}}{\text{Finetuned Model Perf.}_{i}}$. If the merged model performs worse than the pretrained model, then model merging loses its purpose. 
% Normalized average performance is utilized to quantify merged model efficacy~\cite{ilharco2022editing,yadav2024ties}. Specifically, for a merged case \(H_{t}^{j}\), we calculate its \emph{normalized average performance} by
% \[
% \frac{1}{t} \sum_{i}^{t} \frac{\text{(Merged Model Perf.)}_{i}}{\text{(Finetuned Model Perf.)}_{i}}
% \]
% Notably, we additionally introduce
% \[
% \text{\emph{lower bound}} = \frac{1}{T} \sum_{i=1}^{T} \frac{\text{Pretrained Model Perf.}_{i}}{\text{Finetuned Model Perf.}_{i}},
% \]
% which serves as a baseline to determine at which point a given model merging method starts to lose value. This would occur when task conflicts are not well-handled, and using the pretrained model actually becomes more favorable. In our case, the \emph{lower bound} is 82.30\%, 88.03\%, and 66.47\% for Flan-T5-base, Flan-T5-large, and Mistral-Instruct, respectively.

\subsection{Performance Comparison}\label{exp_results}
\begin{figure}[t]
    \centering
    \includegraphics[width=0.99\linewidth]{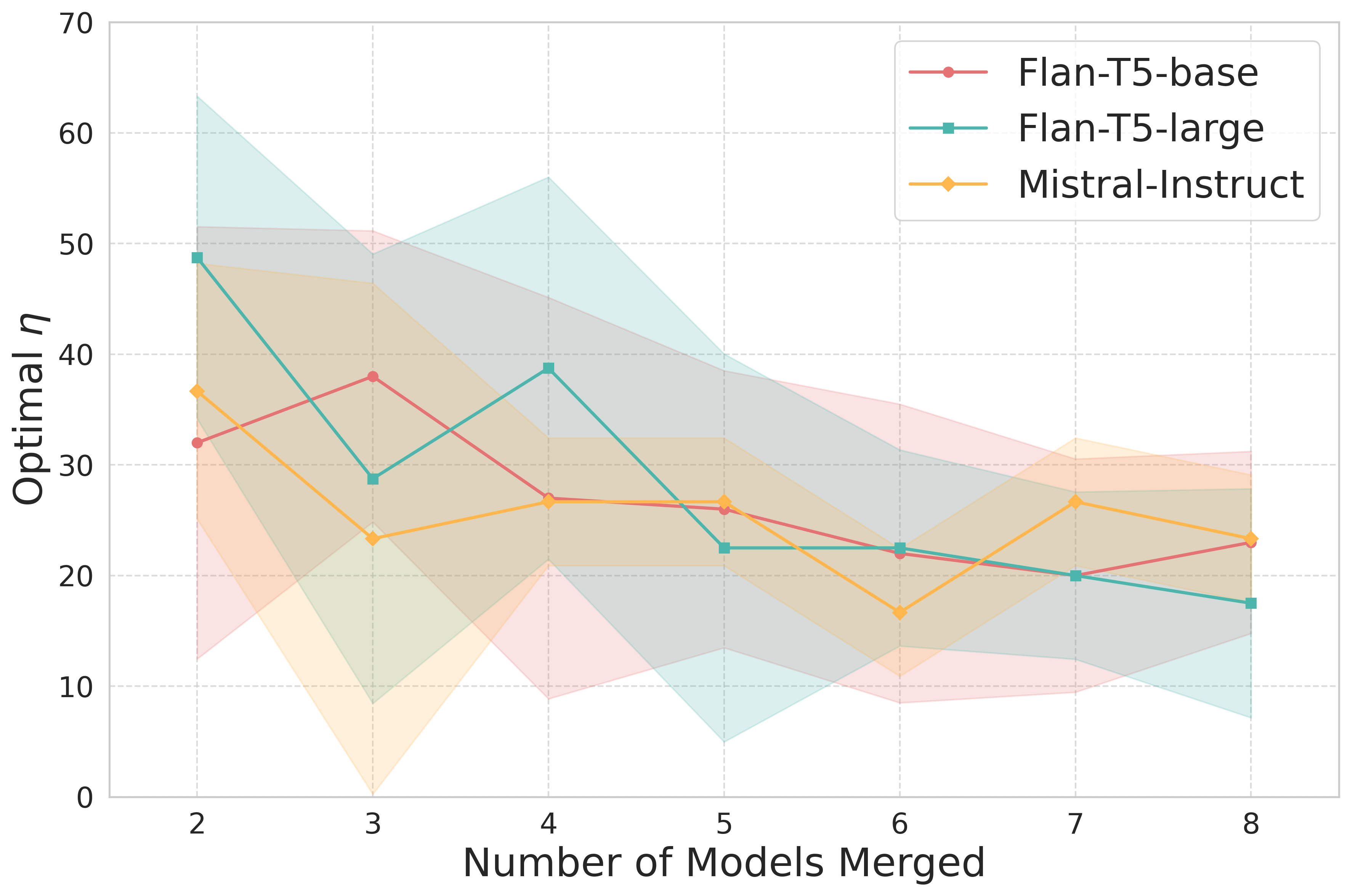}
    \caption{The mean and standard deviation of the optimal \( \eta \), which yields the best merged model performance, decrease as the number of merged models increases.} 
    \vspace{-1em}
    \label{fig:optimal_eta}
\end{figure}
We compare STAR to other data-free approaches, including TIES~\cite{yadav2024ties}, TALL-masks~\cite{wang2024localizing}, which we apply on top of Task Arithmetic~\cite{ilharco2022editing}, i.e., Consensus Task Arithmetic (without tuning the data-dependent hyperparameter \( \lambda_{t} \)), and MetaGPT~\cite{zhou2024metagpt}. Due to the page limit, we defer the discussion around EMR-Merging~~\cite{huang2024emr} and DARE~\cite{yu2024language} to appendix Sec.~\ref{discuss_EMR_Merging} and Sec.~\ref{discuss_DARE}.

The results on Flan-T5-large and Mistral-7B-Instruct are shown in Fig.~\ref{fig:main_datafree} and Flan-T5-base in Fig.~\ref{fig:numerial_evidence}. We note that similar trends as Fig.~\ref{fig:numerial_evidence} can be seen in Fig.~\ref{fig:main_datafree} where the averaged normalized performance decreases as the number of models merged increases, with STAR's performance decay being the slowest across models. On Flan-T5-base, MetaGPT tends to fail quickly, echoing with the findings in~\cite{zhou2024metagpt} - 
% As discussed by~\citet{zhou2024metagpt}, 
MetaGPT may face limitations when merging models of smaller sizes (e.g. Flan-T5-base has only 0.25B parameters) due to its reliance on NTK linearization. 
% We observe similar results here, where MetaGPT tends to fail quickly on the Flan-series models. STAR, however, significantly outperforms the baselines across different model size settings. 
% Up to 20 models were merged in this case, with MetaGPT and STAR show potential to merge even more models without knowledge collapse.
To examine the full potential of each algorithm, we also perform grid search for TIES and STAR and report the best result in Appendix Sec.~\ref{subsec:oneshot_star_is_better}.

\subsection{Additional Results}

\noindent\textbf{Ablation studies on restoring the nuclear norm}
%In appendix Sec.~\ref{subsec:ablation_rescale}, 
In Table~\ref{tab:rescale_help}, we give an example of merging 4 fine-tuned Flan-T5-large models with and without rescale to restore the matrix nuclear norm.
%as introduced in Sec.~\ref{subsec:rescale}. 
%From the table, 
We see that rescale is crucial especially when we use low-rank approximations (e.g. rank-2). 
%1
\begin{table}[t]
\centering
\resizebox{\columnwidth}{!}{ 
\ra{1.8}
\LARGE
\begin{tabular}{@{}ccccccc@{}}\toprule
Rank Kept & Rescale & MRPC & Finance & HellaSwag & PIQA & Avg. Normalized \\ \midrule
r=2 & No & 73.36 & 91.19 & 77.75 & \textbf{80.75} & 97.17 \\
    & Yes & \textbf{74.05} & \textbf{96.04} & \textbf{79.40} & 80.25 & \textbf{99.01} \\ \cmidrule(lr){1-7}
r=4 & No & 73.27 & 94.71 & 78.35 & \textbf{81.00} & 98.32 \\
    & Yes & \textbf{73.79} & \textbf{96.04} & \textbf{79.20} & 80.75 & \textbf{99.02} \\ \cmidrule(lr){1-7}
r=8 & No & 73.44 & 94.71 & 78.70 & \textbf{81.00} & 98.48 \\
    & Yes & 73.44 & \textbf{95.59} & \textbf{78.80} & 80.50 & \textbf{98.58} \\ \cmidrule(lr){1-7}
r=12 & No & 73.44 & 94.71 & 78.55 & \textbf{81.00} & 98.44 \\
    & Yes & 73.44 & \textbf{95.15} & \textbf{78.85} & \textbf{81.25} & \textbf{98.72} \\ \bottomrule
\end{tabular}
}
\caption{The ablation study of the rescaling step to restore nuclear norms (i.e. Sec.~\ref{subsec:rescale}).}
\label{tab:rescale_help}
\end{table}

\begin{figure}[t]
\begin{subfigure}{0.48\textwidth}
\includegraphics[width=0.99\linewidth]{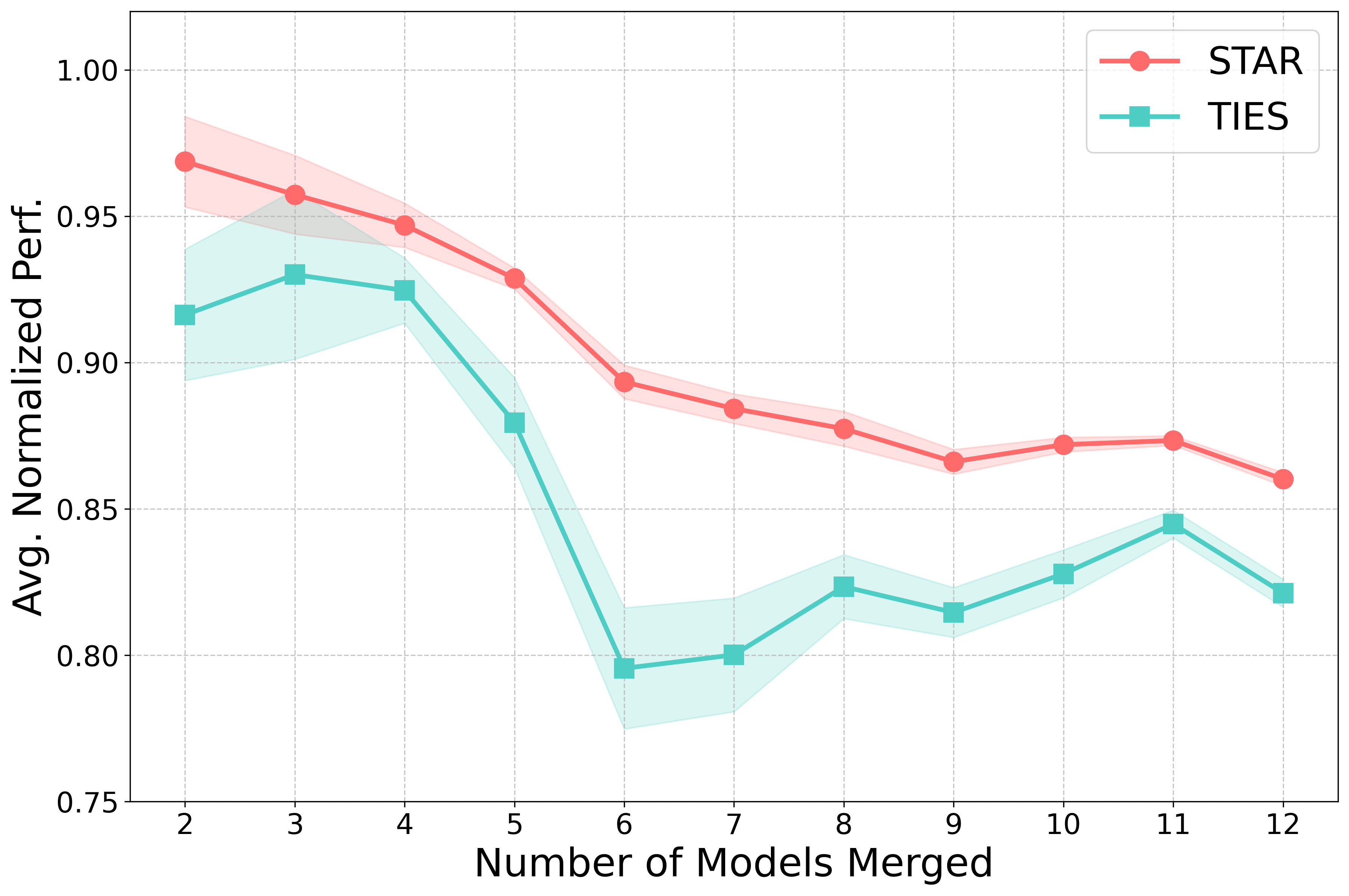} 
\caption{Flan-T5-base}
\end{subfigure}
\begin{subfigure}{0.48\textwidth}
\includegraphics[width=0.99\linewidth]{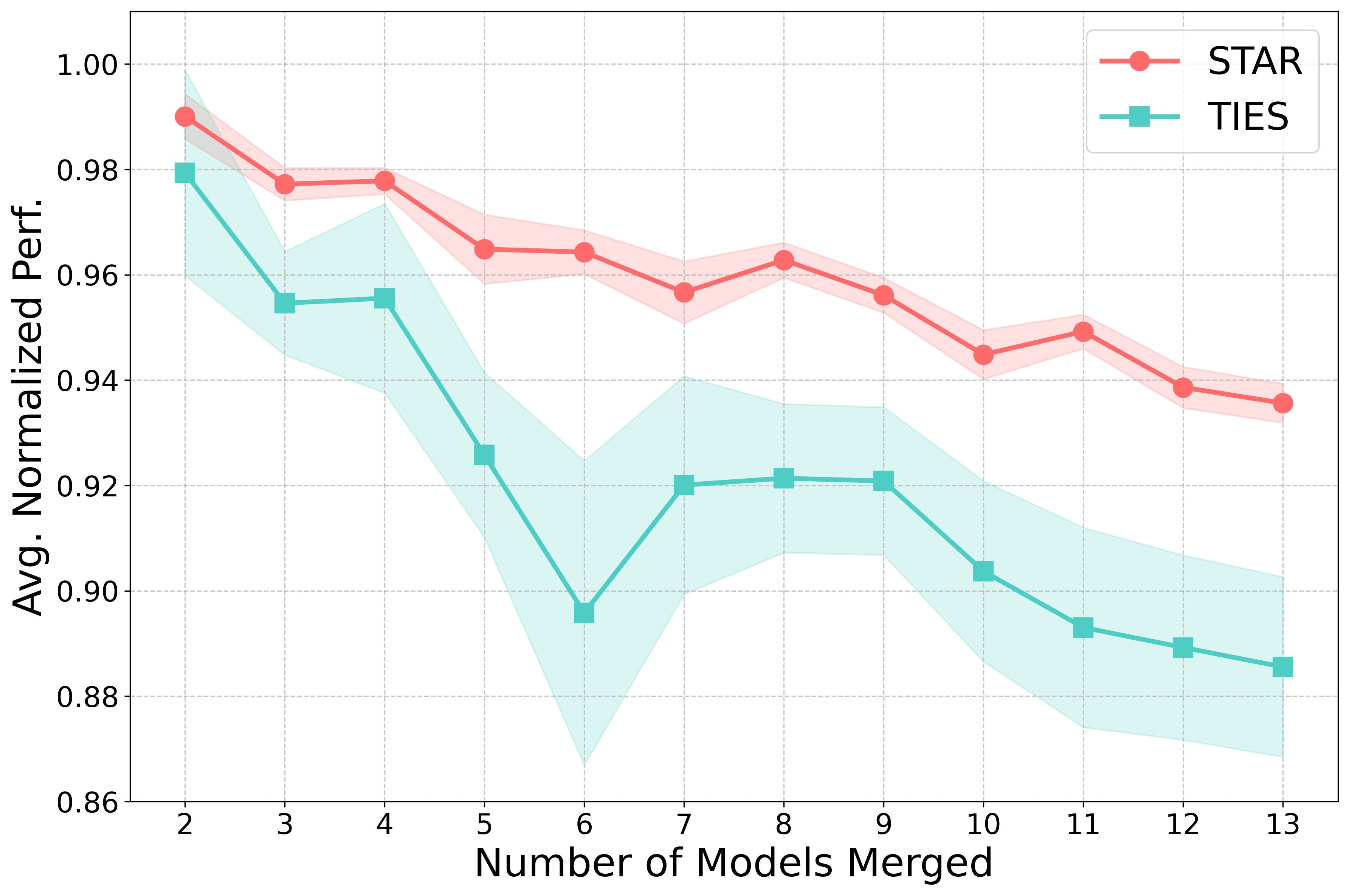}
\caption{Flan-T5-large}
\end{subfigure}
\caption{The average model merging results on Flan-T5-base and Flan-T5-large over a range of possible hyperparameter choices.}
\label{fig:eta_analysis1}
\end{figure}

\noindent\textbf{Sensitivity analysis of $\eta$.} As \(\eta\) is the only tunable hyperparameter in STAR, we further show in 
%appendix 
Fig.~\ref{fig:eta_analysis1} that \(\eta\) is robust across different model merging combinations and numbers of models merged, compared to the baseline (e.g. TIES).
%In this section, 
Specifically, we allow STAR to choose $\eta$ from $\{10, 20, \dots, 70\}$ and TIES to choose $K$ from $\{1, 5, 10, 20, \dots, 70\}$. From the standard deviation in Fig.~\ref{fig:eta_analysis1}, it can indeed be seen that STAR is not sensitive to $\eta$, sparing users' need to fine-tune $\eta$ during the deployment.

\noindent\textbf{Optimal $\eta$ varies as number of models merged.} Following~\citet{ilharco2022editing}, we report the optimal $\eta$ when merging different number of models in Fig.~\ref{fig:optimal_eta}. By searching for \( \eta \) within \(\{10, 20, \dots, 70\}\) across all sampled model merging combinations, we observed an interesting trend: as the number of merged models increases, the optimal \( \eta \) gradually decreases, indicating that higher truncation for each task vector is necessary. 
% This trend is intuitive—task conflicts become more pronounced with more merged models, likely necessitating higher truncation for each task vector. Conversely, when merging fewer models, excessive truncation may be unnecessary, as it could result in a loss of task-specific information.

% \noindent\textbf{More discussions.} Due to the page limit, we put ... in appendix.
% we defer additional analysis on our automatic rank determination schemes in appendix Sec.~\ref{subsec:automatic_rank}. 
% \IKB{Add more references to the appendix once its done.}

\section{Conclusion}
In this paper, we propose Spectral Truncation And Rescale (STAR) for model merging by removing noisy components via spectral decomposition and restoring the original nuclear norm through rescaling. STAR requires no additional inference and is robust to different hyperparameter choices and language models. STAR provides a principaled way of automatic rank determination and is intuitively complimentary to other merging methods.
%% Conclusion
%Future Work
% Global TARE-SV, Knolwedge distribution across layers, matrices....


\section*{Limitation}

While STAR demonstrates strong potential for practical model merging use cases across domains, its performance has been tested primarily on parameter-efficient fine-tuned (PEFT) models in NLP. 
Additionally, STAR requires SVD to orthogonalize task vectors, which may introduce additional computational cost. However, users can mitigate this by leveraging fast SVD algorithms in the implementation.

\section*{Acknowledgement}
This work was primarily done during Yu-Ang Lee's visit to IBM Research, and was supported in part by the National Science and Technology Council, Taiwan, under grant NSTC 113-2628-E-001 -003 -MY4.

\bibliography{custom}

\newpage
\cleardoublepage
\appendix

\section{Appendix}
\subsection{Bounding $\|Bx\|$}
\label{subsec:bound}
Let $r^A$ and $r^B$ be the original ranks of $A$ and $B$, $B=\sum_{i=1}^{r^B}\sigma_i^B u_i^B (v_i^B)^T$, $x=\sum_{j=1}^{r^A} \alpha_j v_j^A$, and $\{v_i^A\}_{i=1}^{r^A}$ and $\{v_i^B\}_{i=1}^{r^B}$ are orthonormal vectors, then we have
\begin{align}
    \|Bx\|&=\|\sum_i\sigma_i^B u_i^B (v_i^B)^T \sum_j \alpha_j v_j^A\| \nonumber\\
    &\leq \sum_i \|u_i^B\| \cdot |\sum_j \sigma_i^B\alpha_j  (v_i^B)^T   v_j^A| \nonumber\\
    &\leq \sum_i \beta\cdot |\sum_j (v_i^B)^T   v_j^A| \nonumber\\
    &\leq \sum_{i=1}^{r^B} \beta \sqrt{r^A} \left(\sum_{j=1}^{r^A} \left(\left(v_i^B\right)^T   v_j^A\right)^2\right)^{1/2} \label{ieq:cauthy}\\
    &= \sum_{i=1}^{r^B} \beta \sqrt{r^A} \left(\sum_{j=1}^{r^A} \left<v_i^B,   v_j^A\right>^2\right)^{1/2},\label{eq:first}
    % \\
    % &\leq r^B \beta \sqrt{r^A},
\end{align}
where $\beta=\max_{i,j}|\sigma_i^B\alpha_j|$, and inequality~\eqref{ieq:cauthy} uses Cauchy-Schwarz inequality. Then we show that
\begin{align}
    1 &= \|v_i^B\|^2 \nonumber\\
    &=\|\sum_{j=1}^{r^A}\left<v_i^B,   v_j^A\right> v_j^A + v_i^{B\perp A}\|^2 \label{eq:decompose}\\
    &=\sum_{j=1}^{r^A} \|\left<v_i^B,   v_j^A\right> v_j^A\|^2 + \|v_i^{B\perp A}\|^2 \label{eq:ptg}\\
    &= \sum_{j=1}^{r^A} \left<v_i^B,   v_j^A\right>^2 + \|v_i^{B\perp A}\|^2 \nonumber\\
    &\geq \sum_{j=1}^{r^A} \left<v_i^B,   v_j^A\right>^2, \label{ieq:second}
\end{align}
where equation~\eqref{eq:decompose} expresses $v_i^B$ by $\{v_i^A\}_{i=1}^{r^A}$, and $v_i^{B\perp A}$ denotes the part of $v_i^B$ that is orthogonal to the span of $\{v_i^A\}_{i=1}^{r^A}$. Equation~\eqref{eq:ptg} follows Pythagorean identity since $v_1^A, v_2^A,\ldots,v_{r_A}^A, v_i^{B\perp A}$ are pairwise-orthogonal vectors. Finally, with Equation~\eqref{eq:first} and~\eqref{ieq:second}, we have
\[
\|Bx\|\leq r^B\beta \sqrt{r^A} .
\]

\newpage
\subsection{Algorithm}

\begin{algorithm}[h]
\caption{Model merging by STAR}\label{alg:STAR}
\begin{algorithmic}
\State \textbf{Input:} \(\bm{\theta}_{\text{pre}}\), \(\{\bm{\theta}_{\text{ft}, i}\}_{i=1}^{T}\), \(\eta\) 
\State \textbf{Output:} \(\bm{\theta}_{\text{merged}}\)
\For{$i = 1$ \textbf{to} $T$} 
    \State $\triangleright$ Get task vector
    \State $\bm{\delta}_{i} \gets \bm{\theta}_{\text{ft}, i} - \bm{\theta}_{\text{pre}}$
    
    \For{$l = 1$ \textbf{to} $L$}
        \State $\triangleright$ SVD
        \State $\bm{u}_{k}, \sigma_{k}, \bm{v}_{k} \gets \textbf{SVD}(\bm{\delta}_{i}^{l})$
        \State $r \gets \textbf{rank\_keep}(\bm{\sigma}, \eta, p)$\label{derived_r}
        \State  $\triangleright$ Rescale Singular Values
        \For{$k = 1$ \textbf{to} $r$}
        \State $\sigma_k^{'} \gets \frac{\|\bm{\sigma}\|_{1}}{\|\bm{\sigma}_{1:r}\|_{1}} \cdot \sigma_k$
        \EndFor
        \State  $\triangleright$ Reconstruct
        \State $\bm{\delta}_{i, \text{out}} \gets \sum_{k=1}^{r} \bm{u}_{k} \sigma_{k}^{'} \bm{v}_{k}$
    \EndFor
\EndFor
\State  $\triangleright$  Simple Averaging
\State $\bm{\delta}_{\text{merged}} \gets \frac{1}{T} \sum_{i=1}^{T} \bm{\delta}_{i, \text{out}}$
\State \textbf{return} $\bm{\theta}_{\text{merged}} \gets \bm{\theta}_{\text{pre}} + \bm{\delta}_{\text{merged}}$
\end{algorithmic}
\end{algorithm}

% \subsection{Ablation studies on retraining the nuclear norm}

% \subsection{Sensitivity Analysis}
% \label{subsec:sensitivity}
% In this section, we allow STAR to choose $\eta$ from $\{10, 20, \dots, 70\}$ and TIES to choose $K$ from $K\in\{1, 5, 10, 20, \dots, 70\}$.
% In Sec.~\ref{subsubsec:oneshot_star_is_better}, we will show that grid search TIES's hyperparameter $K$ helps to improve the performance, but it still performs worse than STAR without grid search over $\eta$. In Sec.~\ref{subsubsec:star_insensitive}, we will further show that STAR is not sensitive to $\eta$, sparing users' need to fine-tune $\eta$ during the deployment.
\subsection{Discussion on EMR-Merging}\label{discuss_EMR_Merging}
EMR-Merging~\cite{huang2024emr} is a recent data-free model merging method that reports outstanding performance with minimal additional storage. It first constructs a unified merged task vector, \( \tau_{\text{uni}} \), which retains the maximum amplitude and sign information shared by all task vectors (\( \tau_i \)). Then, task-specific masks (\( M_i \)) and rescalers (\( \lambda_i \)) are derived based on sign agreement and parameter magnitude alignment between \( \tau_i \) and \( \tau_{\text{uni}} \). Finally, during inference, EMR-Merging dynamically adapts \( \tau_{\text{uni}} \) for each task using 
\[
\hat{W_t} = W_{\text{pre}} + \hat{\tau}_t,
\]
where 
\[
\hat{\tau}_t = \lambda_t \cdot M_t \odot \tau_{\text{uni}}.
\]

In other words, EMR-Merging adjusts model weights at run-time, whereas our approach, along with the included baselines (i.e., TIES, MetaGPT, and TALL-masks), operates statically. This makes direct comparison infeasible; therefore, we do not include EMR-Merging as one of the baselines.

\subsection{Discussion on DARE}\label{discuss_DARE}

STAR follows a similar protocol to DARE~\cite{yu2024language}, as both methods involve two steps: dropping certain components and rescaling. However, there are key differences between them.

On one hand, DARE randomly drops entries of task vectors in parameter space, following:
\[
\mathbf{m}^t \sim \text{Bernoulli}(p),
\]
\[
\tilde{\delta}^t = (1 - \mathbf{m}^t) \odot \delta^t.
\]
In contrast, STAR selectively removes redundant dimensions in spectral space.

On the other hand, DARE's rescaling scheme is based on:
\[
\hat{\delta}^t = \frac{\tilde{\delta}^t}{1 - p},
\]
aiming at approximating the original embeddings, while STAR's rescaling focus on restore the spectral-truncated weight matrices to their original scale.

Unlike STAR, which can function as a standalone model merging method, DARE primarily serves as a plug-in to enhance other merging techniques. For comparison, we follow DARE's protocol and report the results of DARE+TA (Task Arithmetic) and DARE+TIES in Table~\ref{tab:dare_comparison}. Specifically, we vary DARE's drop rate \( p \) from \{0.1, 0.2, \dots, 0.9\}, and the results suggest that even when DARE is applied on top of TA and TIES, STAR still achieves superior performance.

\begin{table}[h]
\centering
\renewcommand{\arraystretch}{1.3} % 增加行距
\resizebox{\linewidth}{!}{ % 自動縮放表格以適應雙欄寬度
\LARGE
\begin{tabular}{lcc}
\toprule
\textbf{Method} & \textbf{Hyperparameter} & \textbf{Avg. Normalized} \\
\midrule
TA & \( \alpha = 0.125 \) & 91.67 \\
TA+\textbf{DARE} & \( \alpha = 0.125, p^* = 0.7 \) & 91.78 \\
TIES & \( k = 20 \) & 93.83 \\
TIES+\textbf{DARE} & \( k = 20, p^* = 0.2 \) & 93.71 \\
STAR & \( \eta = 40 \) & \textbf{95.30} \\
\bottomrule
\end{tabular}
}
\caption{Results from merging eight fine-tuned Flan-T5-large models. TA is fixed with a scaling factor of \( \alpha = 0.125 \), and TIES is set with \( k = 20 \), using the best-performing DARE drop rate (\( p^* \)).}
\label{tab:dare_comparison}
\end{table}

\subsection{One-shot STAR performs even better than grid-search TIES}
\label{subsec:oneshot_star_is_better}
\begin{figure}[h]
\begin{subfigure}{0.48\textwidth}
\includegraphics[width=0.98\linewidth]{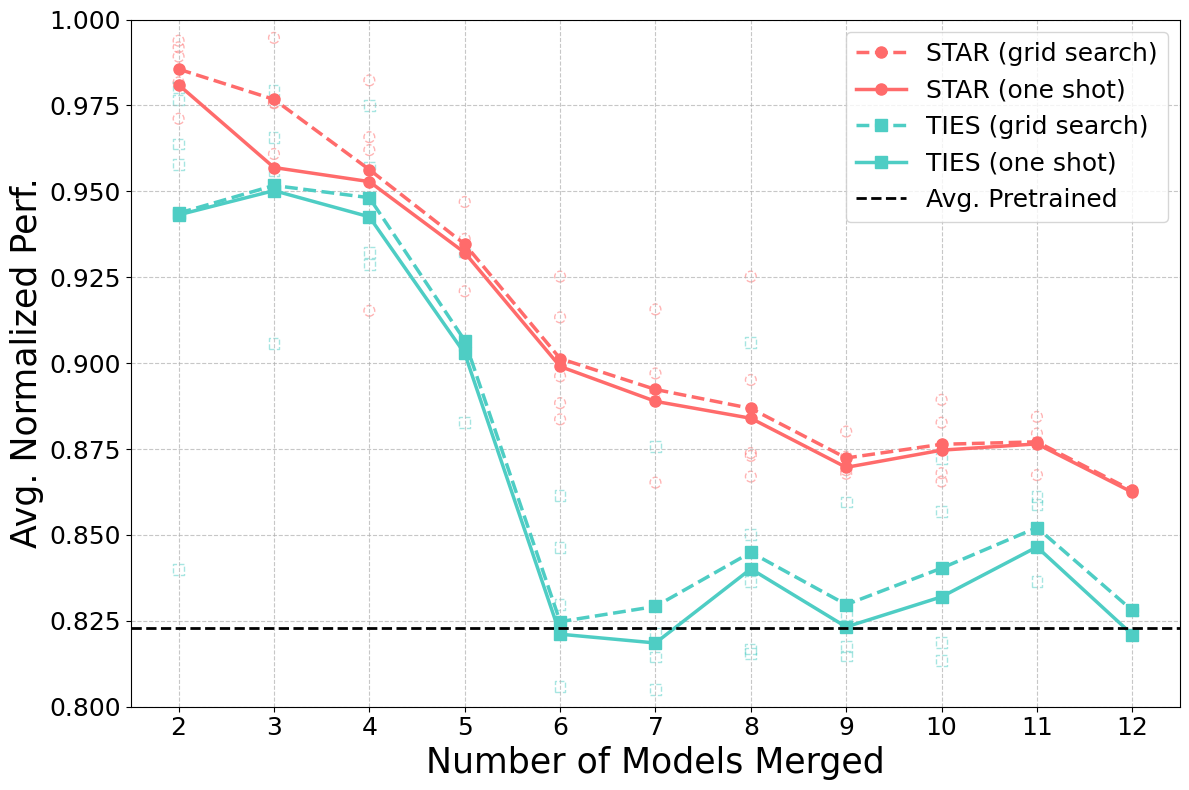} 
\caption{Flan-T5-base}
\label{fig:enter-label}
\end{subfigure}
\begin{subfigure}{0.48\textwidth}
\includegraphics[width=0.98\linewidth]{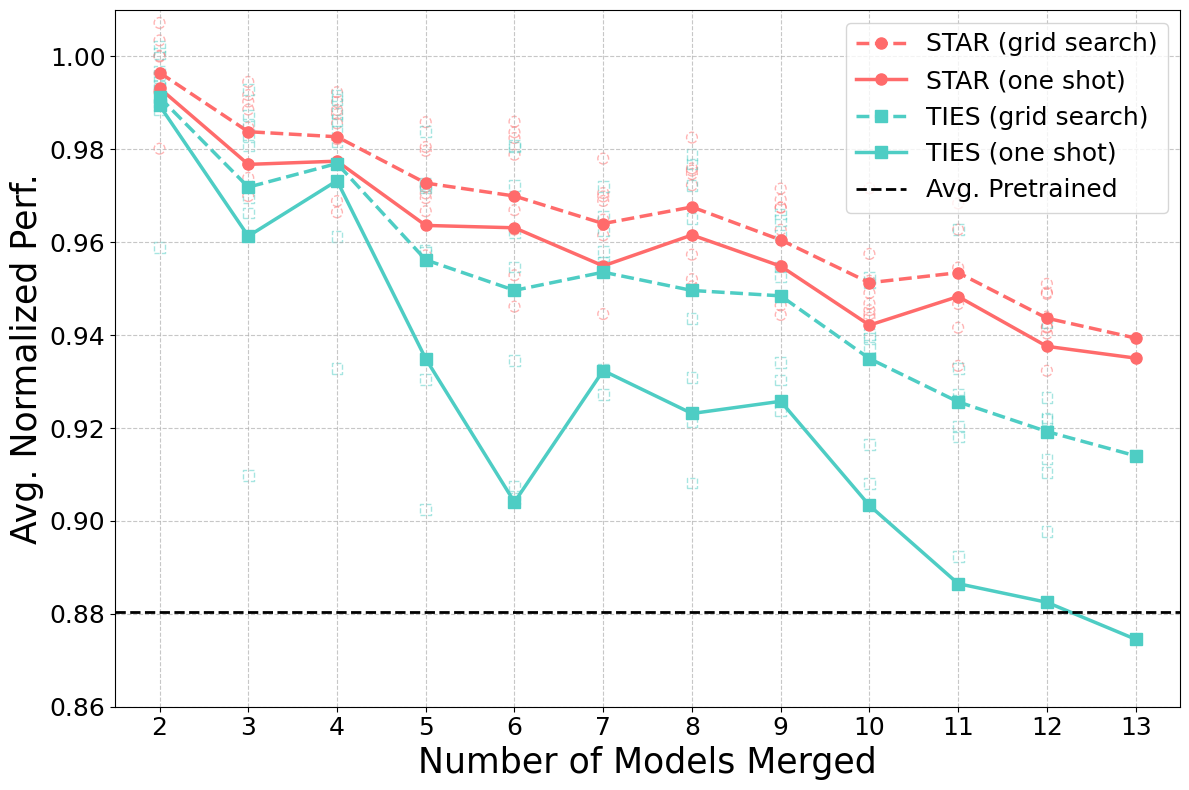}
\caption{Flan-T5-large}
\end{subfigure}
\caption{The model merging results on Flan-T5-base and Flan-T5-large with both pre-determined hyperparameter (one-shot, solid lines) and grid-searched hyperparameter (dashed Lines). The performance of each sampled combinations is represented by shaded dots.
% \IKB{1. Change the marker to distinguish STAR and TIES; 2. Use the same style for one-shot results as in Fig 3, and change grid search lines to dash lines}
}
\label{fig:main_grid}
\end{figure}

Recall that in Fig.~\ref{fig:main_datafree}, we have shown the one-shot performance with pre-determined $K=20$ and $\eta=40$ for TIES and STAR, respectively. In Fig.~\ref{fig:main_grid}, we further show their best possible results over the grids we searched for. Specifically, from Fig.~\ref{fig:main_grid}, we see that the grid search does not improve the performance much on Flan-T5-base for both TIES and STAR. Even after performing grid search for TIES, it still fails to surpass the one-shot performance of STAR, further emphasizing the practicality of our method in real-world applications. On Flan-T5-large, the gain from grid search on TIES becomes obvious especially when we are merging more models. With STAR, grid search over $\eta$ also helps but the results are relatively consistent.

% \newpage

% To take a closer look at the impact of $\eta$ to STAR's performance, we calculate the average performance of STAR with $\eta\in\{10, 20, \dots, 70\}$ and include the standard deviation in Fig.~\ref{fig:eta_analysis1}. On both Flan-T5-base and Flan-T5-large, the standard deviation on STAR's average normalized performance is small, implying STAR is not sensitive to $\eta$ and hence does not require active hyperparameter tuning to perform well.

% \subsection{Single Task Generalization Ability}
% % like model breadcrumbs, do single task performance gain by merging with others!
% % Would be easy(? just pick  out the combination that have avg normalized perf. > 1

\subsection{Details about the fine-tuned models considered in the experiments}\label{app:detail}
\label{subsec:experiment_details}
For Flan-T5-base, we selected 7 LoRA-16 finetuned models from FusionBench\footnote{\url{https://huggingface.co/collections/tanganke}}~\cite{tang2024fusionbench}, which is a benchmark targeted for model merging (excluding only CoLA as it tends to output the same answer), and finetuned 5 additional models ourselves on the Finance, IMDB, AG News, HellaSwag, and BoolQ datasets. We applied the same rank (16) and scaling factor (32) as in FusionBench, with the learning rate and number of epochs tuned on the validation set. Following a similar approach, we selected 7 Flan-T5-large models from FusionBench and finetuned 6 additional models ourselves, including Finance, IMDB, AG News, HellaSwag, and BoolQ, and PIQA.

For Mistral-Instruct, 20 models are selected from the Lots of LoRA collection~\footnote{\url{https://huggingface.co/Lots-of-LoRAs}}~\cite{bruel2024compress}, which encompasses up to 500 diverse task types, making it an ideal environment for evaluating model merging methods.
% , with the selection criteria being ease of evaluation while covering a diverse range of types. 
The considered task IDs are: 039, 190, 247, 280, 290, 298, 330, 357, 363, 391, 513, 564, 587, 834, 846, 1198, 1341, 1391, 1448, 1605.
% \TP{specify the datasets on which these models were trained on. Also include the model ids for both lots of loras and this in the appendix so that others can reproduce it.}\AR{Noted}

% \subsection{An example of the automatic rank and scale determination in STAR}
% \label{subsec:automatic_rank}
% \begin{figure*}[b!]
%     \centering\hspace{8cm}
%     \includegraphics[width=0.90\linewidth]{image/adaptively/piqa_flan_large.png}
%     \caption{An example of the rank automatically determined by STAR using PIQA's task vector on Flan-T5-large. }
%     \label{fig:rank_viz}
% % \end{figure*}

% % \begin{figure*}
%     \centering\hspace{8cm}
%     \includegraphics[width=0.90\linewidth]{image/adaptively/piqa_flan_large_rescale_factor_3.png}
%     \caption{An example of the rescaling factor automatically determined by STAR using PIQA's task vector on Flan-T5-large. 
%     % \textbf{Rescale factor} when \(\eta\) is set to 40\% for PIQA's task vector on Flan-T5-large. Each matrix requires different rescaling factors to restore the nuclear norm, even though the threshold \(\eta\) is applied uniformly. This demonstrates how STAR leverages heterogeneity in sigular values across matrice to accomplish weighted autonomous weighting.
% }

%     \label{fig:rescale_viz}
% \end{figure*}
% By definition, STAR can adaptively determine the number of dimensions needed (rank) to express each weight matrices as well as the scaling factor for each matrices. We give an example of the ranks and scaling factors using PIQA's task vector on Flan-T5-large in Fig.~\ref{fig:rank_viz} and Fig.~\ref{fig:rescale_viz}.

\end{document}